\def\year{2022}\relax
\documentclass[letterpaper]{article} 
\usepackage{aaai22}  
\usepackage{times}  
\usepackage{helvet}  
\usepackage{courier}  
\usepackage[hyphens]{url}  
\usepackage{graphicx} 
\usepackage{gensymb}
\usepackage{booktabs}
\usepackage{caption} 
\DeclareCaptionStyle{ruled}{labelfont=normalfont,labelsep=colon,strut=off} 
\usepackage{algorithm}
\usepackage{algorithmic}

\usepackage{newfloat}
\usepackage{listings}
\floatstyle{ruled}
\newfloat{listing}{tb}{lst}{}
\floatname{listing}{Listing}
\title{Widespread increases in future wildfire risk to global forest carbon offset projects revealed by explainable AI}
\author{
    Tristan Ballard,
    Matthew Cooper,
    Chris Lowrie,
    Gopal Erinjippurath
}
\affiliations{
    Sust Global\\

    San Francisco, California\\
}

\begin{document}

\maketitle

\begin{abstract}
Carbon offset programs are critical in the fight against climate change. One emerging threat to the long-term stability and viability of forest carbon offset projects is wildfires, which can release large amounts of carbon and limit the efficacy of associated offsetting credits. However, analysis of wildfire risk to forest carbon projects is challenging because existing models for forecasting long-term fire risk are limited in predictive accuracy. Therefore, we propose an explainable artificial intelligence (XAI) model trained on 7 million global satellite wildfire observations. Validation results suggest substantial potential for high resolution, enhanced accuracy projections of global wildfire risk, and the model outperforms the U.S. National Center for Atmospheric Research's leading fire model. Applied to a collection of 190 global forest carbon projects, we find that fire exposure is projected to increase 55\% [37-76\%] by 2080 under a mid-range scenario (SSP2-4.5). Our results indicate the large wildfire carbon project damages seen in the past decade are likely to become more frequent as forests become hotter and drier. In response, we hope the model can support wildfire managers, policymakers, and carbon market analysts to preemptively quantify and mitigate long-term permanence risks to forest carbon projects.

\end{abstract}

\section{Introduction}

The United Nations considers carbon offsets and associated carbon credits a critical tool to speed up climate action \cite{UNEP2019}. One of the main types of offset programs is forest carbon offsets, which involve activities such as reforestation, avoidance of deforestation, and improved forest management. In exchange for these actions, forest projects are awarded carbon credits, which can be sold on carbon markets to individuals, businesses, and governments wishing to support the reduction of greenhouse gas emissions. Such market-based mechanisms are designed to incentivize climate action and reduce costs, with one estimate that carbon offset markets can reduce the cost of implementing countries’ current reduction commitments by more than half (\$250 billion) by 2030 \cite{Edmonds2019}. 

Along with the growth in carbon markets is the recognition that carbon offset credits need to be externally validated to determine whether they are credible and will deliver the climate benefits they promise. Central to carbon credits is the idea of permanence, or the long-term stability of emissions reductions claimed by a project, a period of several decades and up to 100 years. This means that the carbon sequestration or emissions reductions associated with the credit must not be reversed in the future, for example by illegal deforestation or natural disturbances. Concern over unverified carbon credits and projects delivering offsets lower than reported has led to claims of ‘greenwashing’. Such credibility doubts can affect individuals and corporations alike. For example, as companies seek to enact their Environmental, Social, and Governance (ESG) criteria, they want to ensure carbon credits purchased to improve their ESG scores are trustworthy.

\begin{figure}[] 
\includegraphics[width=8.5cm]{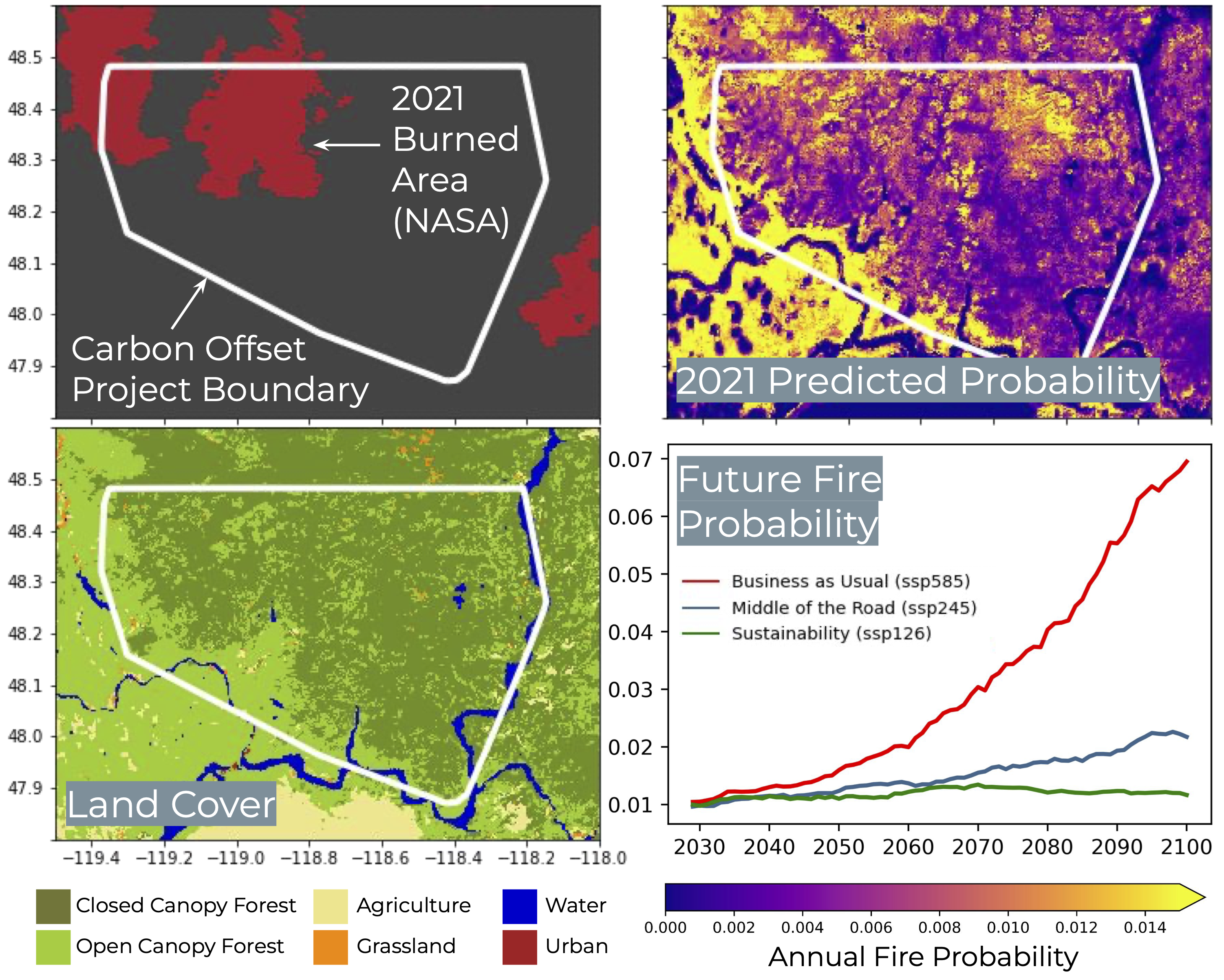}
\centering
\caption{Out-of-sample application of our wildfire model to a forest carbon project (ACR255) in Washington state. A wildfire burned 11\% of the project in 2021 as determined by satellite observations (top left). The 2021 predicted probability (top right) incorporates a range of geospatial features, including land cover (bottom left). Future predictions from our model (bottom right) for this offset project indicate substantial increases in wildfire risks across three candidate climate scenarios (Methods). Project boundaries coarsened for visualization.}
\label{fig1}
\end{figure}

Climate change, through its impacts on wildfire, poses a growing threat to forest carbon projects, jeopardizing their long-term ability to reduce carbon emissions. Wildfires are a present danger to offset projects because they immediately release aboveground carbon to the atmosphere and can likewise increase soil carbon losses \cite{Mekonnen2022}. Unfortunately, recent work suggests current wildfire risk to carbon credits is severely underestimated. A 2022 audit estimated that nearly all of the carbon credits set aside by California’s cap and trade program were exhausted by recent wildfires, a buffer pool intended to last 100 years \cite{Badgley2022}. 

Compounding the effects of wildfires, increased drought stress in response to climate change can limit the ability of forests to sequester carbon and make them more vulnerable to wildfires \cite{Earles2014}. Drought-stressed forests may also be more susceptible to insect infestation, and the range of invasive species is increasing as climates shift \cite{Hicke2020}. For instance, the bark beetle has killed nearly 5\% of western U.S. forests since 1997, and they are expected to continue to spread because of climate change \cite{Hicke2020}. Combined, the interplay of increasing wildfire risk, drought severity, and insect stress in response to climate change can create positive feedback mechanisms that amplify carbon reversal risk in forest carbon projects \cite{Halofsky2020}.

Recent advances in AI can help us better quantify future wildfire risk and the expected long-term impacts on forest carbon projects. Super-resolution neural networks have been built to increase the spatial resolution of fire models, with the AI-enhanced projections used by multiple commercial and nonprofit stakeholders \cite{Ballard2020}. Moreover, Cooper 2022 \cite{Cooper2022}, henceforth ‘Cooper 2022’, created a neural network constrained by scientific understanding of wildfire behavior to improve fire forecasts globally. 

Here we propose a comprehensive update to the Cooper 2022 model, with several methodological advances to improve prediction performance as well as a novel validation against a leading wildfire model. Further, we validate its performance on a large collection of forest carbon offset projects and assess current and future wildfire risks to forest carbon credits.

\section{Data}
Wildfires are a natural part of ecosystems but are increasingly influenced by climate change and human behavior \cite{Abatzoglou2016}. We therefore incorporate a range of high resolution natural and anthropogenic input features indicative of fire initiation and intensity. These inputs include local weather, bioclimatic zones, land cover, topographic characteristics, and population features.

\subsection{Carbon Offset Projects}
We focus on global forest carbon offset projects that are part of the voluntary carbon market and whose credits are eligible for use under the California cap-and-trade program. A total of 190 projects and corresponding polygon boundaries were sourced from the American Carbon Registry, Climate Action Reserve, and Verra project registries (Fig. 2) \cite{So2022}. Projects range in size from less than 1km\textsuperscript{2} to 11,500km\textsuperscript{2}. The largest project, VCS 1566, is associated with 25 million carbon credits \cite{So2022}. A carbon credit is worth 1 tCO\textsubscript{2}e and while varying in value, in 2021 was priced around \$16 per credit \cite{CARB2022}. Project locations span 32 countries, with the most projects in the U.S. (76), China (25), Brazil (20), and Colombia (15) (Fig. 2). Within the U.S., projects span 21 U.S. states, with roughly 1 in 4 in California. 

\subsection{NASA satellite fire occurrence}
We use daily fire data provided by the National Aeronautics and Space Administration (NASA) based on imagery from NASA’s Terra and Aqua satellites \cite{Giglio2009}. NASA’s global fire data product indicates whether fire was observed at a given pixel at a 500m resolution. Because intentional agricultural burning is also detected by the satellites, we constrain model training and validation to non-agricultural areas, isolating for wildfire impacts. Figure 1 shows a satellite detected burn scar from 2021 that burned approximately 11\% of one of the carbon offset projects (ACR255).

\begin{figure}[] 
\includegraphics[width=8.5cm]{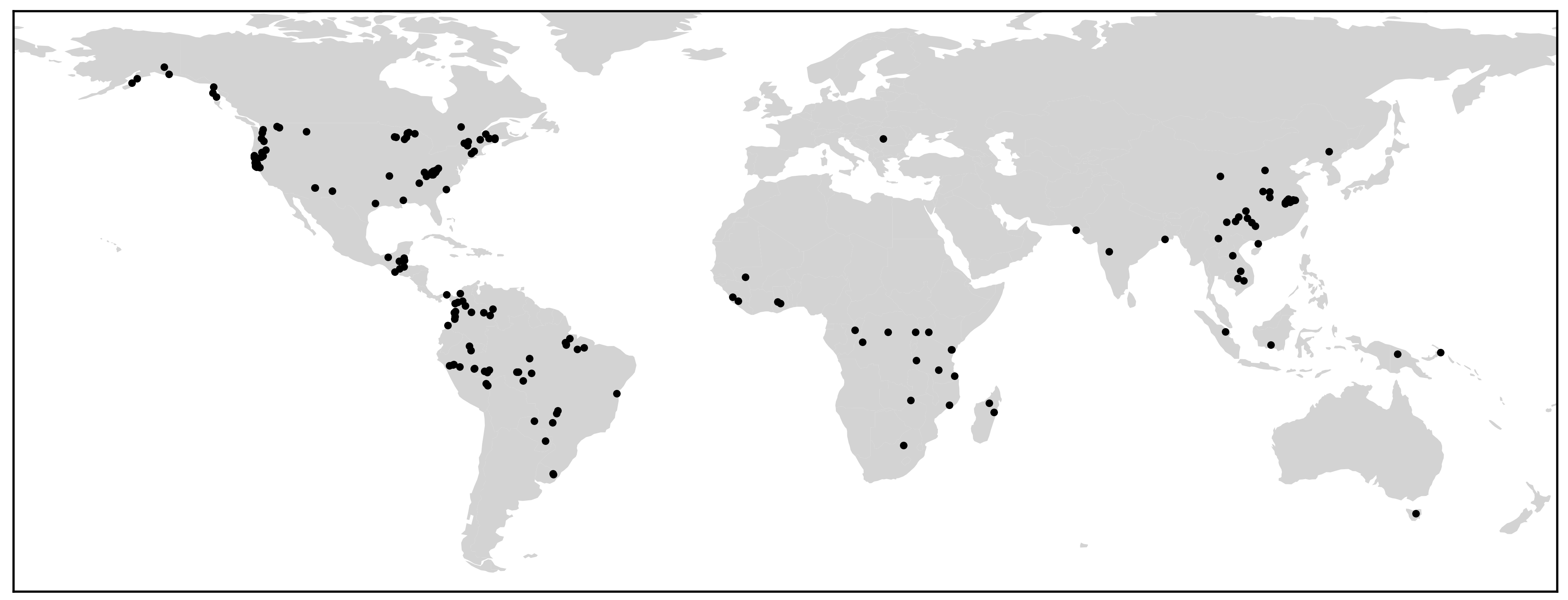}
\centering
\caption{The 190 forest carbon projects evaluated for wildfire risk provide broad geographic coverage, spanning 32 countries and 21 U.S. states.}
\label{fig2a}
\end{figure}

\subsection{Fire Weather}
Because local meteorological conditions are key drivers of wildfires, we derive a daily fire weather index known as the Keetch-Byram Drought Index (KBDI) as a principal input to the model. KBDI is a widely used tool to inform fire management decisions, such as fire suppression strategies \cite{Brown2021}. KBDI is calculated based on daily precipitation and maximum temperature data, with higher values indicating lower soil moisture, influencing the likelihood of ignition and spread of wildfire. We derive historical KBDI values from ERA5-Land data \cite{Munoz2021} available at 10km resolution and future KBDI values from climate model simulations. 

\begin{figure*}[t] 
\includegraphics[width=0.9\textwidth]{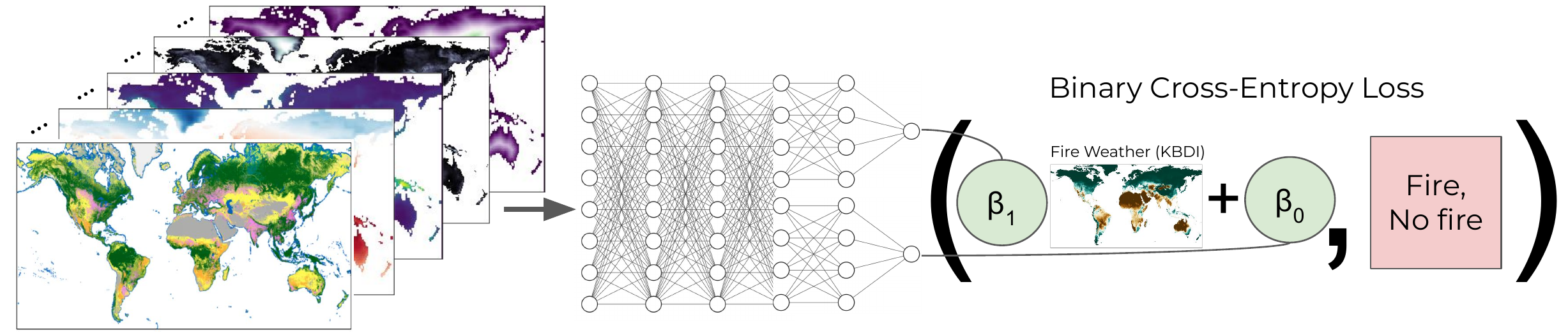}
\centering
\caption{The wildfire model neural network takes as input 55 geospatial weather, bioclimatic zones, land cover, topographic characteristics, and population features to predict wildfire probability. The network combines dense layers with a final constrained, explainable fire weather (KBDI) layer. The penultimate layer nodes serve as interpretable intercept and scaling terms for the KBDI input.}
\label{fig2}
\end{figure*}

\subsection{Climate Change Risk}
We assess future impacts of climate change on wildfires by deriving daily KBDI from climate model simulations of temperature and precipitation. Simulations are from a set of 28 bias-corrected Coupled Model Intercomparison Project phase 6 (CMIP6) models \cite{Thrasher2022}. Recent advances in contrastive learning can also be used to bias-correct and super-resolve the CMIP6 model inputs of future climate \cite{Ballard2022} and is an area of active research. We use both historical simulations spanning 2000 to 2015 and future simulations spanning 2016 to 2100 from three different carbon emissions scenarios. Briefly, the three scenarios are: “Fossil-fueled Development (Business as Usual)” SSP5-8.5, the scenario most consistent with current carbon emissions; “Middle of the Road” SSP2-4.5; and “Sustainability” SSP1-2.6, the scenario consistent with the Paris Climate Accord goal of below 2°C global warming. Subjectively, we view the Middle of the Road scenario, or alternatively a pathway between it and the Business as Usual scenario, as the most realistic.

\subsection{Bioclimatic Zones, Land Cover, and Topography}
Wildfire behavior varies considerably between different ecosystems. For example, small, annual fires are common in savanna ecosystems, while fires in dense North American forests typically have larger footprints but are less frequent. To capture the diversity of bioclimatic zones, we include 19 climate features from WorldClim \cite{Fick2017} often used for ecosystem and species distribution modeling. 

We also include land cover features to complement the bioclimatic zone features. Satellite-derived land cover is provided by the European Space Agency at 300m resolution \cite{ESA2017}. We group the land cover into 6 classes: closed canopy forest, open canopy forest, grassland, urban, agriculture, and Other (e.g. bare soil, ice) (Fig. 1). Through feature engineering experimentation, we found that wildfire probability at a given pixel is correlated not only with its land cover class but also the land cover of its surroundings. We therefore include additional features that summarize the land cover type in a surrounding 500m, 1km, and 2km radius of each pixel. 

We include 6 features corresponding to local topography, such as elevation and slope, which can play a large influence on the spread and suppression of wildfire \cite{TerrainBase}. 

\subsection{Human Dynamics}
Human activity plays a considerable role in modern wildfire behavior. For example, in California, the majority of fires are now started by humans rather than by lightning \cite{Balch2017}. We include 3 features indicative of human activity that may influence fire ignition, suppression, and management. These include subnational GDP as well as each pixel’s distance to cities and accessibility \cite{Kummu2018}. Accessibility, or ‘friction surface’, is a distance measure that accounts for transport networks, providing information on remote geographies difficult for firefighters to reach \cite{Weiss2015}.

\section{Methodology}
The wildfire model is first trained and validated globally on land pixels with and without wildfire. We then further validate model performance on forest carbon project locations before modeling future wildfire risks to the offset projects.

17 million locations were sampled globally at random to form the training and validation sets, with the training and validation sets consisting of 7 million and 10 million locations, respectively. Oceans and water bodies are excluded from sampling, as well as agricultural pixels. We exclude agricultural pixels due to the overwhelming presence of agricultural burning within the observational data. We found excluding these agricultural pixels results in a model more targeted to wildfire prediction rather than intentional burns. Training and validation samples are drawn from 2001 to 2009 and 2010 to 2021, respectively. Each sample includes the annual presence or absence of wildfire as derived from satellites in the sample year (response variable) along with the corresponding observed annual mean KBDI value and time-static features for the location. 

Sample sizes were ultimately determined by computational constraints. The model was trained for 12 hours on 4 NVIDIA Tesla K80 GPUs with 104GB memory. The 7 million training samples represent less than 5\% of the possible training data, suggesting opportunity for further prediction skill improvement. 

\subsection{Network Architecture}
We identified three key design goals for our network architecture:
\begin{itemize}
  \item Explainable AI feature layer for KBDI
  \item Monotonically increasing wildfire probability with KBDI
  \item Nonlinear relationships between remaining features
\end{itemize}

First, the geospatial relationships between KBDI and wildfire probability are of regular interest scientifically and for fire management, so a layer that facilitates KBDI interpretability in particular is desirable. A monotonic relationship between wildfire probability and KBDI is desirable because of the body of scientific literature supporting soil moisture association with wildfire ignition and spread. Such science-informed constraints also have the beneficial effect of reducing the search space, leading to faster training. Last, we expect the relationships between input features to be highly nonlinear at a global level and therefore dense layers within a neural network are desirable. 

To achieve these design goals, we built a network combining feed-forward dense layers with a final KBDI layer with the desired constraints (Fig. 3). All 54 input features aside from KBDI are fed into dense layers with two final output nodes. These nodes are then fed into a final layer where one node serves as an intercept term $\beta_0$ and the other as a scaling term $\beta_1$ for the KBDI input. The single output of this final layer serves as a probability estimate for the binary response variable [fire, no fire], with binary cross-entropy used as the loss function. 

The KBDI layer with constraints achieves the goal of ensuring monotonicity and interpretability. While we do not strictly enforce a monotonically increasing relationship, as expected scientifically (drier conditions associated with higher wildfire risk), in practice this is the result. The intercept term $\beta_0$ (Fig. 3) can be interpreted as baseline global wildfire risk, while the scaling term $\beta_1$ (Fig. 3) reflects the sensitivity of a location to increases in fire weather like warming and drought. Thus, in a location with a high value for $\beta_0$ and a low value for $\beta_1$, there is a high historical probability of wildfire occurrence, but wildfire probability is relatively less sensitive to weather and future climate impacts than changes in other features in the model. Conversely, in a location with a low value for $\beta_0$ and high value for $\beta_1$, fire probability has historically been low, but the model expects wildfire probability to be highly sensitive to novel future conditions  under climate change. We save a presentation of this relationship, along with its variation geographically and in time, for a separate analysis, since our focus here is on carbon project exposure. 

More details of the model architecture and typical out-of-sample performance can be found in Cooper 2022. While the architectures are similar, several advancements have been made to improve performance: Removal of intentional agricultural burns (see Methods); Inclusion of topographic features; Doubling the total number of input features; Annual aggregation of daily time-varying variables to align with stakeholder use cases; and Modifications to the number of nodes and layer connections. We elect not to publish precise hyperparameter choices like number of nodes per layer, learning rate scheduler, and activation functions for this model, though we highlight such information is publicly available for Cooper 2022.

\begin{figure}[] 
\includegraphics[width=8.5cm]{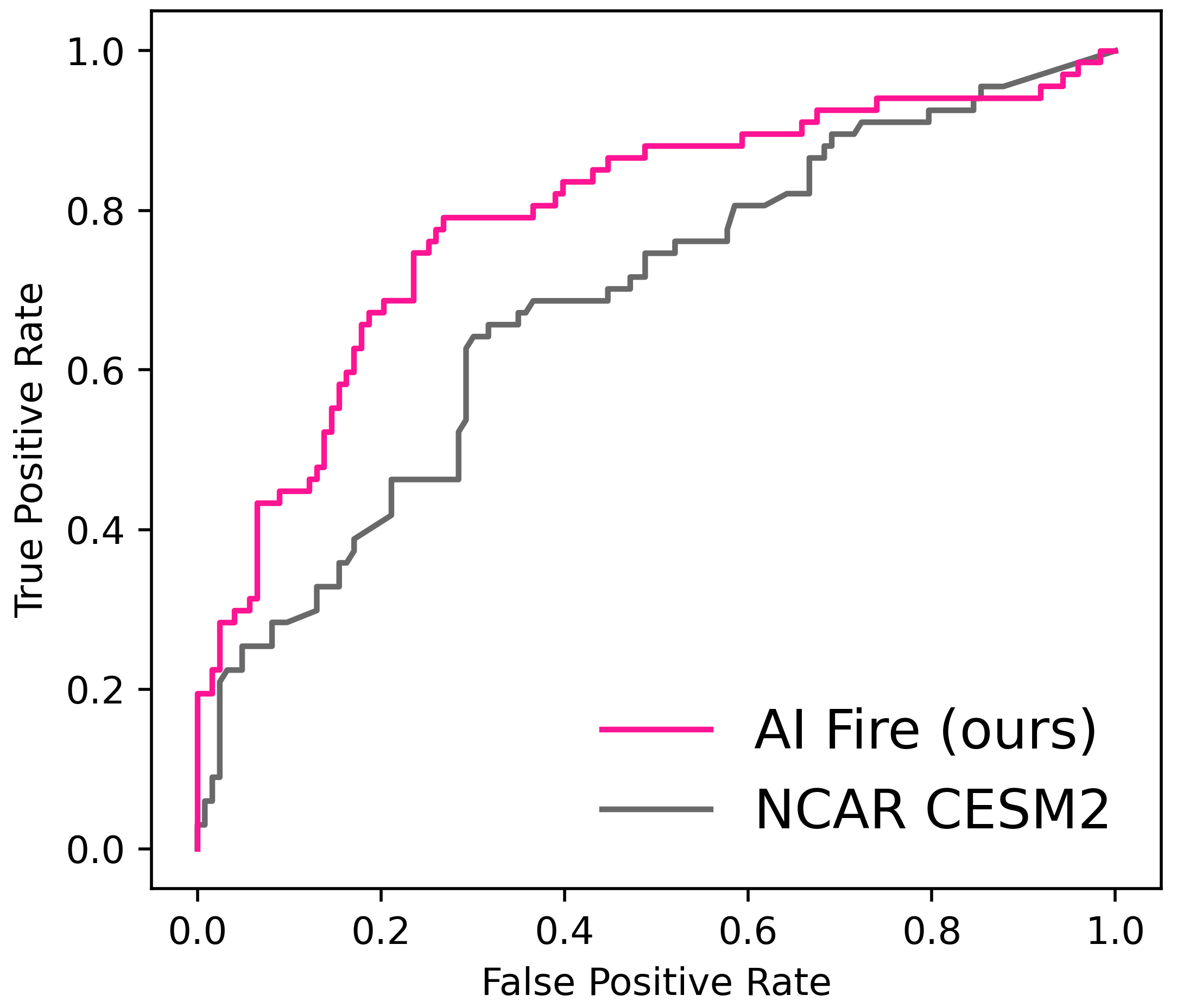}
\centering
\caption{To support application to forest carbon projects, we validate the wildfire model (pink) on the global projects during the held out validation period from 2010-2021. We compare Area under the Curve (AUC) scores against NCAR's leading climate model, CESM2 (grey). The wildfire model outperforms the benchmark CESM2 model. The wildfire model similarly outperforms Cooper 2022 (Appendix).}
\label{fig3}
\end{figure}

\subsection{Validation}
First, we evaluate predictions on the 10 million global held out sample points with sample years corresponding to 2010 to 2021. In short, we find that this model outperforms Cooper 2022 and the benchmark models referenced therein (not shown). This improvement in performance supports the model’s credibility in predicting global wildfire occurrence. 

Second, because we are interested in the model’s prediction skill in particular regions, namely the carbon offset projects, we evaluate its performance locally in these regions before making any projections of future carbon offset project risk. To further support application of the model to the carbon offset project collection, we benchmark predicted wildfire probabilities against projections from a leading fire simulation model from CMIP6, the National Center for Atmospheric Research’s (NCAR) Community Earth System Model 2 (CESM2) model \cite{Danabasoglu2020}. CESM2 is a state-of-the-art global climate model that includes a fire model estimating burned area, and we use the ensemble mean burned area from 8 runs of the climate model averaged over the validation period 2011 to 2021. Like our model, the CESM2 fire model incorporates features related to dryness, land cover, and human dynamics. A primary contrast between the two models is that our model is largely data-driven, while the CESM2 model is process-based. Ours is also capable of forecasting at far higher spatial resolution, 300m as opposed to 100km from CESM2.

\section{Results}
Wildfire was observed at 67 of 190 global offset projects during the 2010 to 2021 validation period, suggesting these forest carbon projects already are exposed to significant wildfire risk. Of those, 39 projects saw more than 10\% of the project burned, representing a substantial impact to the carbon sequestration potential of these forest projects. 18 additional projects experienced wildfire from 2001 to 2009, highlighting that even if projects did not experience wildfire during the validation period, they may still be at risk.

Evaluating the wildfire model performance on the forest carbon offset projects, we find that the wildfire model substantially outperforms the benchmark NCAR CESM2 model. To assess performance, we compare satellite observed wildfires within the carbon project boundaries during the held out validation period (Fig. 1), and calculate Area under the Curve (AUC) scores for each model (Fig. 4). Higher AUC scores are indicative of better predicting models. The wildfire model has an AUC score of 0.79, and the benchmark CESM2 model has an AUC score of 0.68 for global projects (Fig. 4). Similar validation over U.S. projects suggests the wildfire model is an improvement relative to the Cooper 2022 model as well (Appendix). Given the superior performance of the wildfire model, we proceed to simulating future wildfire risk using CMIP6 climate simulations of temperature and precipitation as inputs.   
 
We find that wildfire risk to carbon offset projects across the U.S. is projected to increase 34\% [CI: 24-46\%] by 2050 and 249\% [195-337\%] by 2080 under the Business as Usual climate scenario (Fig. 4). Confidence intervals in brackets represent climate model input uncertainties only, so these are conservative estimates of uncertainty. Increased wildfire risk by the end of the century is even more striking, at 376\% [251-771\%] under the same scenario. This ‘runaway’ wildfire risk increase is likely driven by the extreme level of global warming projected under this scenario, although combined increases in severe drought can also increase wildfire risk. For context, the Business as Usual climate scenario was designed by climate scientists and policymakers as a worst case pathway of unchecked emissions, with many optimistic that it is avoidable if countries adhere to current commitments.

Wildfire risk in 2050 and beyond is markedly reduced under the more optimistic Middle of the Road climate scenario, though risk is still expected to increase by 21\% [17-31\%] by 2050 and 55\% [37-76\%] by 2080 (Fig. 5). Even under the Sustainability pathway, risk increases 18\% [15-19\%] by 2050 and remains at similarly elevated levels through the end of the century. This suggests that regardless of future reductions in greenhouse carbon emissions, considerable increases in wildfire risk are already ‘locked in’, posing a threat to carbon credit permanence. 

\begin{figure}[] 
\includegraphics[width=8.5cm]{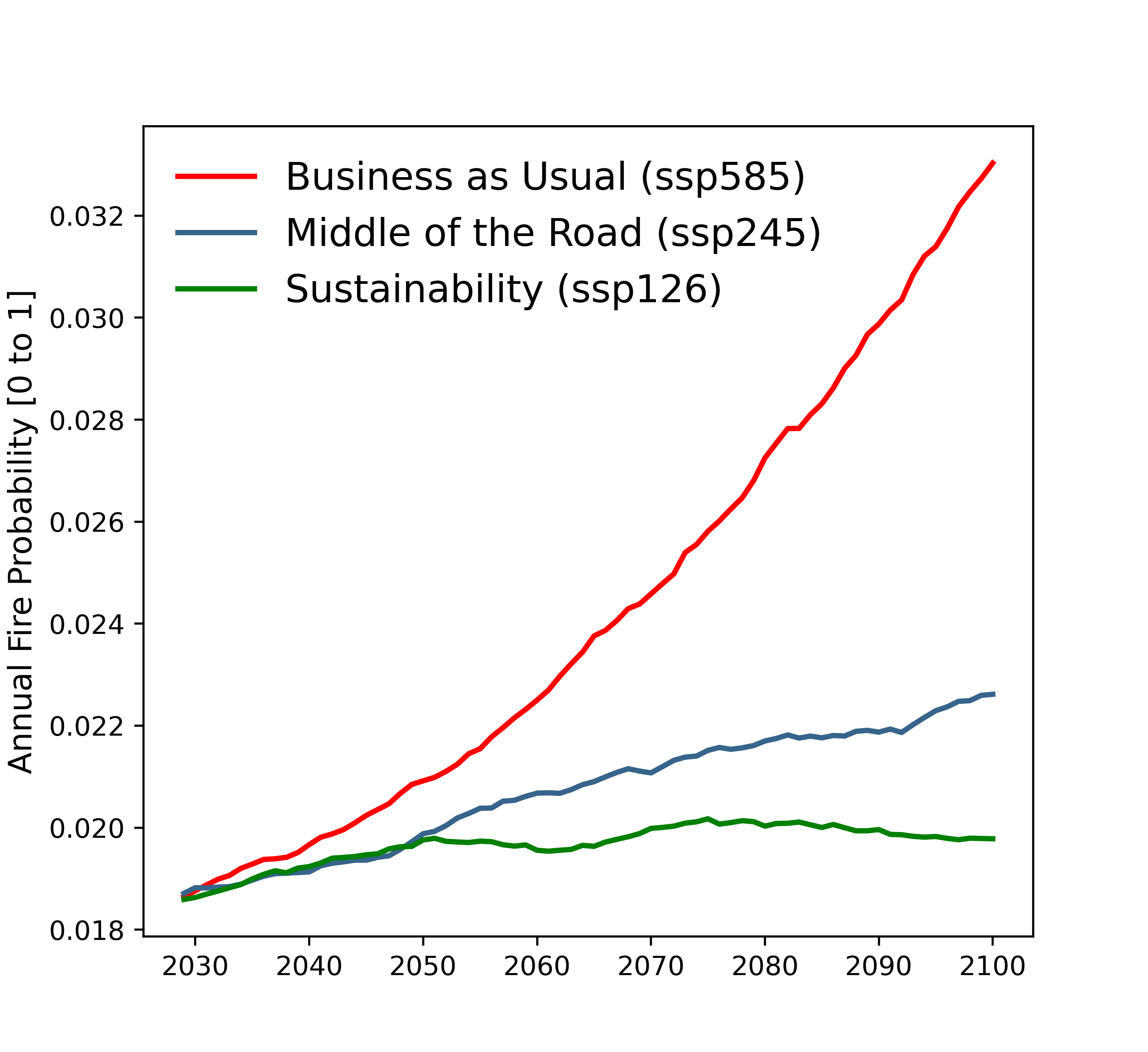}
\centering
\caption{Forest carbon projects face substantial increases in wildfire probability in the coming decades. To simulate future wildfire probability, the wildfire model was paired with climate inputs from an ensemble of 28 CMIP6 models under three distinct climate emissions scenarios. Each line is the mean estimated probability across the 190 offset projects, with a 10-year smoothing applied for visualization.}
\label{fig4}
\end{figure}

\section{Conclusion}
 
Forest carbon credits are intended to be a long-term solution to mitigating climate change, and buyers want to ensure that the credits they purchase will provide lasting benefits to the environment. However, we find that nearly half (45\%) of global forest carbon offset projects examined have experienced at least one wildfire since satellite records began in 2001. These projects thus already experience high levels of fire exposure, jeopardizing their permanence and long-term stability. 

Here we propose an explainable artificial intelligence (XAI) model of wildfire risk trained on millions of global locations and incorporating a range of geospatial features. Validation results for the wildfire model suggest considerable potential for high resolution, enhanced accuracy projections of global wildfire risk. The wildfire model outperforms the U.S. National Center for Atmospheric Research's leading wildfire model as well as Cooper 2022, a prior model version that previously underwent global validation and is actively used by various commercial and nonprofit partners. The higher fidelity simulations of present and future wildfire risk made possible by our wildfire model can support climate scientists and a broad range of stakeholders alike.

We find that the baseline carbon offset wildfire exposure observed in this study is projected to increase substantially, upwards of 55\% [37-76\%] by 2080 under a mid-range scenario, a clear concern for carbon credit permanence and tackling climate change. Already, wildfires within offset projects in the past decade alone have exhausted nearly all of the carbon credits that California’s cap and trade program set aside for wildfire losses, and that reserve was intended to last 100 years \cite{Badgley2022}. Unfortunately, our results indicate such large carbon credit losses of the past decade are likely to become far more frequent in the coming decades as forests become hotter and drier. An encouraging outcome of our results is that they show significant reductions in wildfire risk can be achieved if countries commit to meeting emissions reduction targets that aim to keep emissions below the Middle of the Road climate scenario. 

A variety of stakeholders can benefit from quantifying climate risks to forest carbon offset projects as illustrated here. Policymakers can set aside a greater share of carbon credits in buffer pools for wildfire losses in order to maintain carbon market stability, since our results indicate the losses of previous years are likely to increase \cite{Buchholz2022}. Wildfire managers can also use the projections to identify projects at risk and take mitigation actions, such as controlled burns. Carbon market analysts and ESG auditors can similarly use this model to assess the permanence and credibility of carbon credits from active and proposed forest offset projects globally.

\section{Future Directions}
While this model represents a clear improvement over the baseline NCAR model and Cooper 2022, the 7 million points used in training represent less than 5\% of available data, so the model will likely continue to improve as more data is included. To improve future forecasts, we also plan to incorporate recent advances \cite{Ballard2022} in climate model generative adversarial networks (GANs) that can be used to bias-correct and super-resolve the CMIP6 climate model inputs for more accurate long-term wildfire forecasts. 

\section{Data Availability}
Future wildfire projections for each forest carbon project are available at https://developers.sustglobal.com.

\bibliographystyle{unsrt}


\setcounter{table}{0}
\setcounter{figure}{0}
\renewcommand{\thetable}{A\arabic{table}}
\renewcommand\thefigure{A\arabic{figure}}    

\section{Appendix}

Carbon Project IDS included in the analysis: 'ACR360', 'ACR255', 'ACR273', 'CAR973', 'ACR274', 'ACR324', 'ACR199', 'CAR1183', 'ACR248', 'ACR211', 'ACR427', 'ACR303', 'CAR1175', 'ACR249', 'CAR1314', 'ACR417', 'CAR1197', 'CAR1215', 'CAR1213', 'ACR247', 'ACR373', 'CAR1205', 'CAR1264', 'ACR276', 'CAR1257', 'ACR267', 'ACR202', 'ACR260', 'ACR280', 'CAR1208', 'CAR1217', 'CAR1191', 'ACR210', 'CAR1013', 'ACR279', 'CAR1041', 'ACR361', 'CAR1095', 'ACR282', 'CAR1209', 'ACR281', 'ACR262', 'CAR1066', 'CAR1180', 'CAR1046', 'CAR1297', 'ACR425', 'ACR284', 'CAR1032', 'CAR1190', 'ACR458', 'CAR993', 'CAR1173', 'ACR371', 'ACR393', 'ACR256', 'ACR173', 'ACR292', 'ACR257', 'ACR192', 'CAR1204', 'CAR1174', 'CAR1104', 'CAR1102', 'VCSOPR10', 'ACR182', 'ACR377', 'CAR1103', 'ACR200', 'ACR378', 'ACR189', 'ACR288', 'CAR1094', 'ACR423', 'VCS1052', 'VCS1094', 'VCS1112', 'VCS1113', 'VCS1118', 'VCS1122', 'VCS1201', 'VCS1202', 'VCS1233', 'VCS1311', 'VCS1317', 'VCS1325', 'VCS1326', 'VCS1327', 'VCS1339', 'VCS1351', 'VCS1359', 'VCS1382', 'VCS1389', 'VCS1390', 'VCS1391', 'VCS1392', 'VCS1395', 'VCS1396', 'VCS1398', 'VCS1399', 'VCS1400', 'VCS1403', 'VCS142', 'VCS1463', 'VCS1477', 'VCS1503', 'VCS1530', 'VCS1538', 'VCS1541', 'VCS1542', 'VCS1558', 'VCS1566', 'VCS1571', 'VCS1650', 'VCS1663', 'VCS1664', 'VCS1674', 'VCS1684', 'VCS1686', 'VCS1689', 'VCS1695', 'VCS1704', 'VCS1740', 'VCS1769', 'VCS1826', 'VCS1882', 'VCS1897', 'VCS1911', 'VCS1935', 'VCS1969', 'VCS2079', 'VCS2082', 'VCS2083', 'VCS2087', 'VCS2249', 'VCS2250', 'VCS2252', 'VCS2278', 'VCS2290', 'VCS2293', 'VCS2310', 'VCS2322', 'VCS2343', 'VCS2345', 'VCS2362', 'VCS2373', 'VCS2375', 'VCS2378', 'VCS2379', 'VCS2386', 'VCS2387', 'VCS2396', 'VCS2397', 'VCS2398', 'VCS2399', 'VCS2401', 'VCS2403', 'VCS2410', 'VCS2451', 'VCS2476', 'VCS2477', 'VCS2481', 'VCS2504', 'VCS2506', 'VCS2507', 'VCS514', 'VCS562', 'VCS576', 'VCS605', 'VCS612', 'VCS673', 'VCS687', 'VCS738', 'VCS799', 'VCS812', 'VCS829', 'VCS832', 'VCS868', 'VCS872', 'VCS875', 'VCS934', 'VCS953', 'VCS958', 'VCS959', 'VCS960', 'VCS963', 'VCS977', 'VCS981', 'VCS985', 'VCS987'.

\begin{figure}[] 
\includegraphics[width=8.5cm]{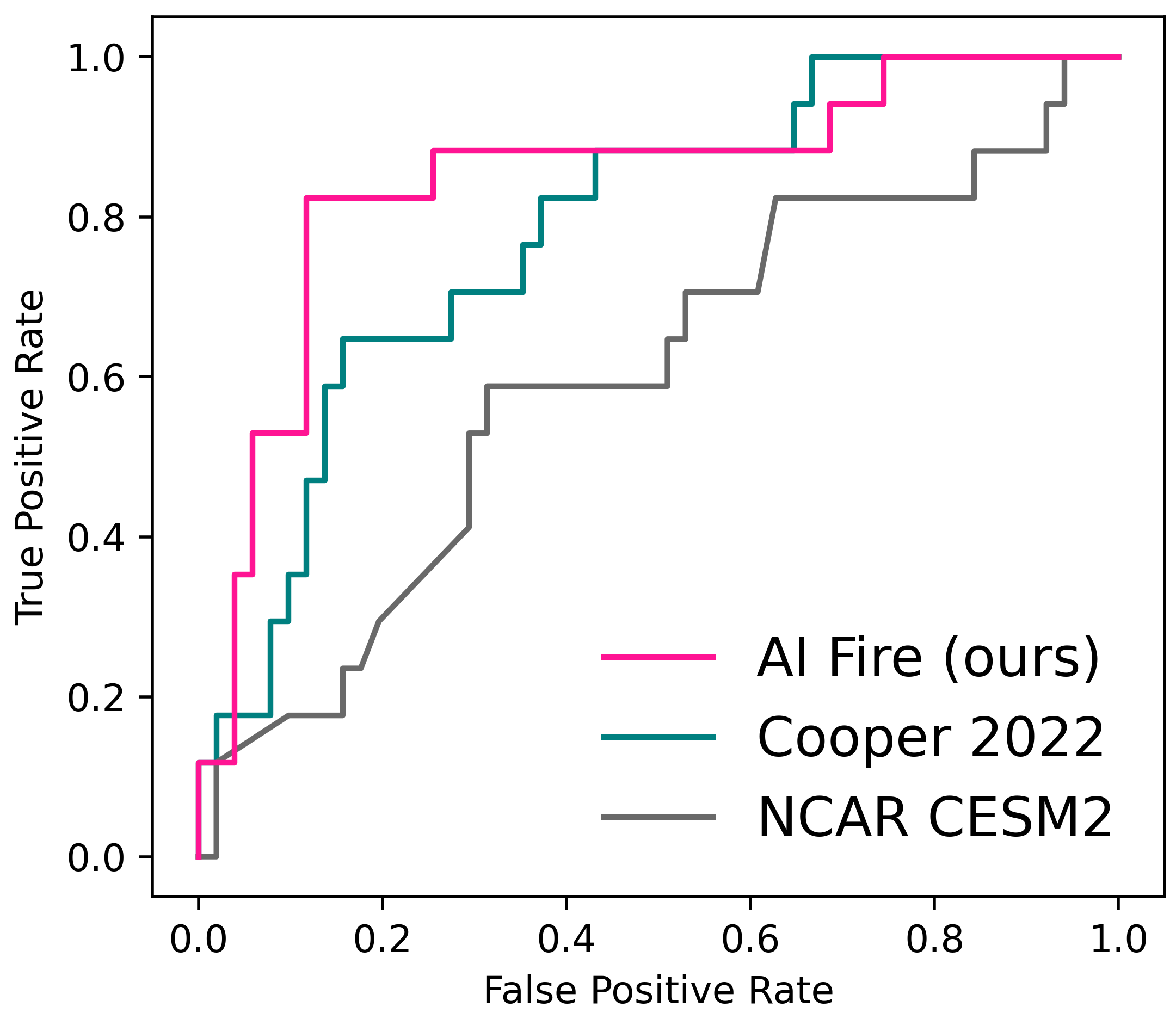}
\centering
\caption{To support application of the fire model to forest carbon projects, we validate the model (pink) during the held out validation period from 2010-2021, here for the subset of 68 contiguous U.S. projects. As in Fig. 4, we compare Area under the Curve (AUC) scores against CESM2 (grey), NCAR’s leading climate model, as well as Cooper 2022 (teal), a prior model version that has previously undergone global validation. The wildfire model (pink) outperforms both benchmark models. The Cooper 2022 model was run only for the contiguous U.S. here to conserve computation resources. For these U.S. projects, the wildfire fire model has an AUC score of 0.85, and the benchmark CESM2 model has an AUC score of 0.60. The Cooper 2022 model has an AUC score of 0.78, suggesting the fire model is an improvement relative to it as well. }
\label{fig30}
\end{figure}

\begin{figure}[] 
\includegraphics[width=8.5cm]{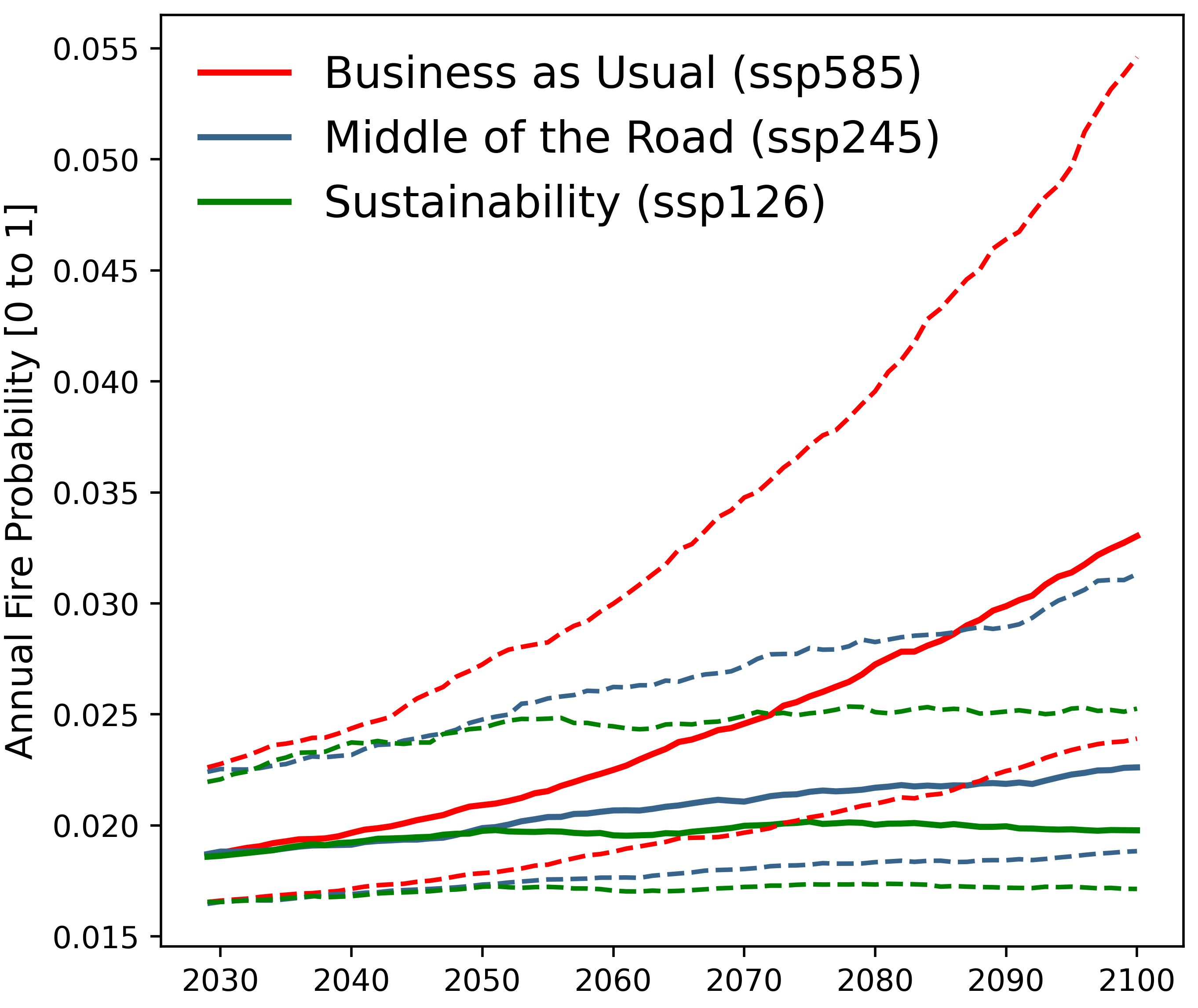}
\centering
\caption{Equivalent to Figure 5, showing uncertainty ranges in dashed lines. Lower and upper bounds reflect the 16th and 84th percentile of the 28 CMIP6 model projections.}
\label{fig30}
\end{figure}

\clearpage
\end{document}